\documentclass[review,hidelinks,onefignum,onetabnum]{siamart250211}

\usepackage{lipsum}
\usepackage{amsfonts}
\usepackage{graphicx}
\usepackage{epstopdf}
\usepackage{algorithmic}
\ifpdf
  \DeclareGraphicsExtensions{.eps,.pdf,.png,.jpg}
\else
  \DeclareGraphicsExtensions{.eps}
\fi

\newsiamremark{remark}{Remark}
\newsiamremark{hypothesis}{Hypothesis}
\crefname{hypothesis}{Hypothesis}{Hypotheses}
\newsiamthm{claim}{Claim}
\newsiamremark{fact}{Fact}
\crefname{fact}{Fact}{Facts}

\headers{Reconstruct multidimensional random field models}{M. Xia and Q. Shen}

\title{Efficient reconstruction of multidimensional random field models with heterogeneous data using stochastic neural networks\thanks{Submitted to the editors 11/17/2025.}}

\author{Mingtao Xia\thanks{Department of Mathematics, University of Houston, Houston, TX 
  (\email{mxia4@uh.edu}, \url{https://sites.google.com/nyu.edu/mingtao-xia/home}).}
\and Qijing Shen\thanks{Nuffield Department of Medicine, University of Oxford, Oxford, UK 
  (\email{qijing.shen@ndm.ox.ac.uk }).}
  }

\usepackage{amsopn}

\ifpdf
\hypersetup{
  pdftitle={An Example Article},
  pdfauthor={M. Xia and Q. Shen}
}
\fi

\usepackage{graphicx} 
\usepackage{amsmath}
\usepackage{amssymb}
\usepackage{url}
\usepackage{bm}
\usepackage{booktabs}
\date{July 2025}
\def\E{\mathbb{E}}
\def\d{\mbox{d}}
\usepackage{mathtools,nccmath}
\usepackage{bm}
\newtheorem{example}{Example}[section]
\usepackage{changes}
\usepackage{comment}
\usepackage{algorithm}
\usepackage{algorithmic}

\begin{document}
\nolinenumbers
\maketitle

\begin{abstract}
    In this paper, we analyze the scalability of a recent Wasserstein-distance approach for training stochastic neural networks (SNNs) to reconstruct multidimensional random field models. We prove a generalization error bound for reconstructing multidimensional random field models on training stochastic neural networks with a limited number of training data. Our results indicate that when noise is heterogeneous across dimensions, the convergence rate of the generalization error may not depend explicitly on the model's dimensionality, partially alleviating the "curse of dimensionality" for learning multidimensional random field models from a finite number of data points. Additionally, we improve the previous Wasserstein-distance SNN training approach and showcase the robustness of the SNN. Through numerical experiments on different multidimensional uncertainty quantification tasks, we show that our Wasserstein-distance approach can successfully train stochastic neural networks to learn multidimensional uncertainty models.
\end{abstract}



\section{Introduction}
\label{sec:introduction}

The accurate quantification of uncertainty is a cornerstone of reliable prediction and robust design across different disciplines. For example, in climate modeling, ensembles of forecasts are used to quantify uncertainty in global temperature projections and extreme weather events~\cite{Palmer2019, Tebaldi2021}. As another example, financial risk assessment relies on stochastic models to estimate the value at risk and the expected shortfall of portfolios under volatile market conditions~\cite{McNeil2015, escobar2011risk}. Additionally, in the field of autonomous systems, uncertainty quantification (UQ) is critical for perception algorithms (e.g., estimating the confidence in object detection) and for enabling safe reinforcement learning policies that can account for unpredictable environments~\cite{Michelmore2020, Thrun2005}. Across these diverse disciplines, researchers have to make critical inferences and decisions based on computational models whose inputs, parameters, and governing equations are imperfectly known and subject to uncertainties~\cite{Sullivan2015}.

Traditional methods for such UQ tasks are predominantly based on utilizing parametric families of distributions, such as multivariate Gaussians or Gaussian copulas~\cite{Xiu2010}. These models are analytically tractable, often requiring only the computation of sample means and covariances. However, when little information is given about the unknown distribution, which does not fall into the parametric families of distributions, those traditional methods may not apply. These limitations are particularly acute in high-dimensional settings, where the curse of dimensionality renders simple parametric models increasingly misrepresentative and can lead to severe underestimation of extreme-event probabilities~\cite{embrechts2013modelling, murphy2023probabilistic}.

More flexible, non-parametric approaches, such as the kernel density estimation method \cite{okabe2009kernel}, offer an alternative, but are themselves susceptible to the ``curse of dimensionality", requiring sample sizes that grow exponentially with dimension~\cite{Wasserman2006}. This has spurred significant interest in developing scalable inference techniques that can operate in high dimensions without imposing overly restrictive assumptions. Recently, the fusion of generative models, such as generative adversarial networks (GANs), with principles from optimal transport (OT) has emerged as a powerful paradigm for learning high-dimensional distributions~\cite{Goodfellow2014}. Other generative modeling approaches, such as the variational encoder method and the normalization flow approach, also find wide applications in UQ \cite{Kingma2013, winkler2019learning}. 
These methods learn a transformation (typically a neural network) that maps a simple base distribution (e.g., a standard Gaussian) to a complex data distribution. 

The OT-based approach minimizes the Wasserstein distance between the generated and empirical distributions, providing a geometrically meaningful and often more stable training objective compared to traditional adversarial losses~\cite{Arjovsky2017}. Recently, a novel Wasserstein distance approach has been proposed for directly training stochastic neural networks (SNNs) by minimizing a \textbf{local squared Wasserstein-2 (}$\bm{W_2}$) \textbf{distance loss function}. This approach has been demonstrated to be more effective than some other machine-learning-based benchmarks, such as WGANs and Bayesian neural networks \cite{xia2024local, xia2025generalized, xia2025new}, in reconstructing a random field model:
\begin{equation}
    \bm{y}_{\bm{x}} = \bm{f}(\bm{x};\omega), \bm{x}\in D\in\mathbb{R}^n
    \label{model_objective}
\end{equation}
using an SNN whose output is:
\begin{equation}
    \hat{\bm{y}}_{\bm{x}} = \hat{\bm{f}}(\bm{x};\hat{\omega}).
    \label{model_objective_approximate}
\end{equation}
In Eqs.~\eqref{model_objective} and \eqref{model_objective_approximate}, $D$ is a bounded set, $\bm{y}_{\bm{x}}, \hat{\bm{y}}_{\bm{x}}\in\mathbb{R}^d$. In Eq.~\eqref{model_objective}, $\omega$ are latent random variables, and $\hat{\omega}$ in Eq.~\eqref{model_objective_approximate} refers to the uncertain weights in the SNN. 
However, the scalability of this approach to reconstruct multidimensional random field models is questionable, as the optimal generalization error bound (expected errors in the testing data) with $N$ training data for this method is $O(N^{-\frac{2}{d}})$ when noise is homogeneous across all dimensions. Therefore, the expected testing error increases fast as the dimensionality of the uncertainty model $d$ increases.

 In this work, we perform an analysis of the scalability of the $W_2$-distance-based approach for learning multidimensional random field models. We investigate the generalization error bound when training SNNs by minimizing the local squared $W_2$ distance loss function for learning multidimensional and high-dimensional random field models. Specifically, our results indicate that the generalization error bound of the trained SNN can be much improved when the noise in the unknown random field model is heterogeneous across different dimensions or the noise lies in a low-dimensional manifold. We also show that the distribution of the SNN is robust against perturbations in the SNN's parameters and improve the previous local squared $W_2$ method to better reconstruct multidimensional uncertainty models in various UQ tasks.

  Our primary contributions are as follows:
\begin{itemize}
    \item[\textbf{1.}] We present an analysis of the generation error bound for reconstructing multidimensional random field models with a finite number of training data. Our results indicate that, when the noise in the model is heterogeneous across directions, the convergence rate might not depend explicitly on the dimensionality of the model. Our result partially alleviates the "curse of dimensionality" when applying the local squared $W_2$ method to reconstruct unknown random field models by training an SNN.
    \item[\textbf{2.}] We improve the recently proposed Wasserstein-distance approach for training SNNs to make it more robust to sparse training data. Theoretically, we proved that the SNN we use is robust against perturbations in the means and variances of its weights, as well as perturbations in its biases. Numerically, we showcase the efficiency of our proposed approach in different multidimensional UQ tasks and demonstrate that it is more accurate in reconstructing random field models like Eq.~\eqref{model_objective} than some benchmark methods.
\end{itemize}

The remainder of this paper is organized as follows. In Section~\ref{generalization}, we present our results on the generalization error bound of learning multidimensional random field models. We propose an improved $W_2$ approach for training SNNs to reconstruct the random field model Eq.~\eqref{model_objective} and carry out numerical experiments to test its efficacy in Section~\ref{numerical_experiments}. In Section~\ref{summary}, we summarize our work and discuss potential future directions. Mathematical proof is given in Appendix~\ref{proof_thm1} and Appendix~\ref{snn_robust}.

\section{Generalization error bound on reconstructing multidimensional random field models}
\label{generalization}
In this section, we analyze the generalization error bound under the $W_2$ distance metric when learning multidimensional random field models with a finite number of training data. Our result gives an improved generalization error bound for training the SNN through minimizing the local (generalized) squared $W_2$ loss function in \cite{xia2024local, xia2025new}. First, we formally define the squared $W_2$ distance between two probability measures that we shall use throughout this manuscript.

\begin{definition}
\rm 
\label{def:W2}
For $\bm{y}_{\bm{x}}, \hat{\bm{y}}_{\bm{x}}\in\mathbb{R}^d$ defined in Eqs.~\eqref{model_objective} and ~\eqref{model_objective_approximate}, we assume that
\begin{equation}
    \E[\|\bm{y}_{\bm{x}}\|^2]< \infty,\,\,\,\,\E[\|\hat{\bm{y}}_{\bm{x}}\|^2]< \infty,\,\,\forall \bm{x}\in D.
\end{equation}
Throughout this manuscript, $\|\cdot\|$ is the $\ell^2$ norm of a vector. We denote the probability measures associated with $\bm{y}_{\bm{x}}$ and $\hat{\bm{y}}_{\bm{x}}$ by $\mu_{\bm{x}}$ and $\hat{\mu}_{\bm{x}}$, respectively. The $\bm{W_2}$ \textbf{distance} between $\mu_{\bm{x}}$ and $\hat{\mu}_{\bm{x}}$ is defined as:
\begin{equation}
W_{2}(\mu_{\bm{x}}, \hat{\mu}_{\bm{x}}) \coloneqq \inf_{{\pi\in \Pi_{\mu_{\bm{x}}, \hat{\mu}_{\bm{x}}}}}
\E_{(\bm{y}_{\bm{x}}, \hat{\bm
{y}}_{\bm{x}})\sim {\pi}}\big[\|{\bm{y}}_{\bm{x}} - \hat{{\bm{y}}}_{\bm{x}}\|^{2}\big]^{\frac{1}{2}}.
\label{pidef}
\end{equation}
In Eq.~\eqref{pidef}, 
\begin{equation}
\begin{aligned}
    \Pi_{\mu_{\bm{x}}, \hat{\mu}_{\bm{x}}} &:= \big\{ \pi \;|\; \pi \text{ is a probability measure defined on~} 
\mathcal{B}(\mathbb{R}^d \times \mathbb{R}^d) \\
&\hspace{2cm} \text{with marginals}~ \mu_{\bm{x}} \text{ and } \hat{\mu}_{\bm{x}} \big\},
\end{aligned}
\end{equation}
{where $\mathcal{B}(\mathbb{R}^d \times \mathbb{R}^d)$ denotes the Borel $\sigma$-algebra associated with $\mathbb{R}^d \times \mathbb{R}^d$.} 
\end{definition}

We shall prove the following result on the convergence rate of using a finite-sample empirical probability measure to approximate the ground truth probability measure under the $W_2$ metric.

\begin{theorem}
\rm
\label{theorem1}
    Let $\mu\in\mathcal{P}(\mathbb{R}^d)$. We assume that:
    \begin{equation}
M\coloneqq\big(\int_{\mathbb{R}^d}\|\bm{y}\|_6^6\mu(\d\bm{y})\big)^{\frac{1}{6}}<\infty,
    \end{equation}
    where $\|\cdot\|_6$ denotes the $\ell^6$ norm of a vector.
    We denote $\sigma_i\coloneqq \big(\int_{\mathbb{R}^d}y_i^6\mu(\d\bm{y})\big)^{\frac{1}{6}}$, and we assume that $\sigma_1\geq\ldots\geq\sigma_d>0$.
    There exists a constant $C$ depending only on $d$ such that, in the $N\rightarrow\infty$ limit,
    \begin{equation}
        \E\left[ W_2^2(\mu_N, \mu) \right] \leq CM^2 \times
        \begin{cases}
            N^{-1/2} \log(1 + N) & \text{if } d\leq 4, \\
            N^{-\frac{1}{2}}+\Big(\prod_{i=1}^d\big(\frac{\sigma_i}{\sigma_1})N\Big)^{-2/d} & \text{if } d>4.
        \end{cases}
        \label{moment_estimate}
    \end{equation}
\end{theorem}

When $d\leq 4$, the proof of Theorem~\ref{theorem1} is the same as the proof of \cite[Theorem 1]{fournier2015rate}. When $d>4$, we prove Theorem~\ref{theorem1} in Appendix~\ref{proof_thm1}. Compared to \cite[Theorem 1]{fournier2015rate}, Theorem~\ref{theorem1} gives a refined convergence rate when using a finite-sample empirical probability measure to approximate the ground truth probability measure in the $W_2$ metric. To exemplify how the factor $\big(\prod_{i=1}^d(\frac{\sigma_i}{\sigma_1})\big)^{-2/d}$ actually gives a faster convergence rate when noise is heterogeneous across different dimensions, we consider the case when there exists a constant $c_0>0$ such that:
\begin{equation}
    \frac{\sigma_i}{\sigma_1}\leq \exp(-c_0i).
    \label{hetero_noise}
\end{equation}
Therefore, we have:
\begin{equation}
    \prod_{i=1}^d\frac{\sigma_i}{\sigma_1}= \exp\big(-c_0\frac{d(d+1)}{2}\big).
\end{equation}
Thus, we have:
\begin{equation}
    \mathbb{E}\big[W_2^2(\mu_N,\mu)\big]\leq C\big(N^{-\frac{1}{2}} + N^{-\frac{2}{d}}\exp(-c_0d)\big).
    \label{decay_rate_property}
\end{equation}
The first term in Eq.~\eqref{decay_rate_property} converges at a rate of $N^{-\frac{1}{2}}$. For the second term, it takes its maximum when $d=(\frac{2\log N}{c_0})^{\frac{1}{2}}$:
\begin{equation}
    N^{-\frac{2}{d}}\exp(-c_0d)\leq N^{-\sqrt{\frac{2c_0}{\log N}}} \exp(-\sqrt{2\log Nc_0}).
    \label{d_maximum}
\end{equation}
 Finally, we have:
\begin{equation}
    \mathbb{E}[W_2^2(\mu_N,\mu)]\leq C\big(N^{-\frac{1}{2}} + N^{-\sqrt{\frac{2c_0}{\log N}}} \exp(-\sqrt{2\log Nc_0})\big),
\end{equation}
and the convergence rate $\big(N^{-\frac{1}{2}} + N^{-\sqrt{\frac{2c_0}{\log N}}} \exp(-\sqrt{2\log Nc_0})\big)$ does not depend on the dimensionality $d$ as $N\rightarrow\infty$.

Next, we consider the 
\textbf{local squared} $\bm{W_2}$ \textbf{loss function} in \cite{xia2024local} for training SNNs defined as:
    \begin{equation}
        W_{2, \delta}^{2, \text{e}}(\bm{y}_{\bm{x}}, \hat{\bm{y}}_{\bm{x}})\coloneqq \int_D W_2^2(\mu^{\text{e}}_{\bm{x},\delta}, \hat{\mu}^{\text{e}}_{\bm{x}, \delta})\nu^{\text{e}}(\d \bm{x}).
    \label{average_w2_local}
    \end{equation}
    In Eq.~\eqref{average_w2_local}, $\mu^{\text{e}}_{\bm{x},\delta}, \hat{\mu}^{\text{e}}_{\bm{x}, \delta}$ are the empirical conditional probability measures associated with $\bm{y}_{\tilde{\bm{x}}}$ and $\hat{\bm{y}}_{\tilde{\bm{x}}}$ conditioned $\|\tilde{\bm{x}}-\bm{x}\|\leq\delta$, and $\nu^{\text{e}}$ is the empirical probability measure of $\bm{x}$.
The local squared $W_2$ loss function is a proxy to the average squared $W_2$ distance between the two random fields $\bm{y}_{\bm{x}}, \hat{\bm{y}}_{\bm{x}}$ in Eqs.~\eqref{model_objective} and \eqref{model_objective_approximate} defined as:
    \begin{equation}
        W_{2}^{2}(\bm{y}_{\bm{x}}, \hat{\bm{y}}_{\bm{x}})\coloneqq \int_D W_2^2(\mu_{\bm{x}}, \hat{\mu}_{\bm{x}})\nu(\d \bm{x}),
    \label{average_w2}
    \end{equation}
    where $\nu$ is the probability measure that each $\bm{x}$ is independently sampled from.
Eq.~\eqref{average_w2} measures how close the approximate random field model Eq.~\eqref{model_objective_approximate} is to the ground truth model Eq.~\eqref{model_objective}.
With a refined convergence rate of the empirical probability measure to the ground truth probability measure in Theorem~\ref{theorem1} for heterogeneous noise, we can improve the generalization error bound (\cite[Theorem 2.2]{xia2025generalized} and \cite[Theorem 4.3]{xia2024local}) on using the generalized local squared $W_2$ loss function Eq.~\eqref{average_w2_local} with a finite number of training data. 
\begin{theorem}
\rm
    \label{theorem_2}
    For every $\bm{x}\in D$, we denote $\mu_{\bm{x}}$ and $\hat{\mu}_{\bm{x}}$ to be the probability measures associated with $\bm{y}_{\bm{x}}$ and $\hat{\bm{y}}_{\bm{x}}$, respectively.
    We make the following assumptions:
    \begin{enumerate}
        \item $\bm{y}_{\bm{x}}$ and $\hat{\bm{y}}_{\bm{x}}$ in Eqs.~\eqref{model_objective} and \eqref{model_objective_approximate} are uniformly bounded for all $\bm{x}\in D$ by a constant $M_0$:
        \begin{equation}
            \|\bm{y}_{\bm{x}}\|<M_0, \,\, \|\hat{\bm{y}}_{\bm{x}}\|<M_0.
            \label{upper_boundM}
        \end{equation}
        \item The probability measures associated with $\bm{y}_{\bm{x}}$ and $\hat{\bm{y}}_{\bm{x}}$ in Eqs.~\eqref{model_objective} and \eqref{model_objective_approximate} are Lipschitz continuous in the $W_2$ distance sense:
    \begin{equation}
        W_2(\mu_{\bm{x}}, \mu_{\tilde{\bm{x}}})\leq L\|\bm{x}-\tilde{\bm{x}}\|,\,\, W_2(\hat{\mu}_{\bm{x}}, \hat{\mu}_{\bm{x}})\leq L\|\bm{x}-\tilde{\bm{x}}\|.
        \label{l_condition}
    \end{equation}
    \end{enumerate}
    For each $\bm{x}\in D$, we denote the number of samples $(\tilde{\bm{x}}, \bm{y}_{\bm{\tilde{x}}})\in S$ such that $\|\tilde{\bm{x}}-\bm{x}\|_2\leq\delta$ to be $N(\bm{x}, \delta)$. 
     We have the following finite-training-sample generalization error bound when using Eq.~\eqref{average_w2_local} to approximate the average squared $W_2$ distance Eq.~\eqref{average_w2}:
    \begin{equation}
    \E\Big[\big|W_2^2(\bm{y}_{\bm{x}}, \hat{\bm{y}}_{\bm{x}}) - W_{2, \delta}^{2, \text{e}}(\bm{y}_{\bm{x}}, \hat{\bm{y}}_{\bm{x}})\big|\Big] \leq \frac{4M_0}{\sqrt{N}} + 8CM_0\E\big[h(N(\bm{x}, \delta), {d})\big] + 8\sqrt{M_0}L\delta,
        \label{theorem2_result}
    \end{equation}
    where $N$ is the number of training samples.
    
    In Eq.~\eqref{theorem2_result}, $C$ is a constant, $N$ is the total number of data points $\{(\bm{x}, \bm{y}_{\bm{x}})\}$, and $L$ is the Lipschitz constant in Eq.~\eqref{l_condition}. In Eq.~\eqref{theorem2_result}, 
    \begin{equation}
h(N, d)\coloneqq\left\{
\begin{aligned}
&2N^{-\frac{1}{4}}\log(1+N)^{\frac{1}{2}}, d\leq4,\\
&2\Big(N^{-\frac{1}{4}}+\big(\prod_{i=1}^d\big(\frac{\sigma_i}{\sigma_1}) N\big)^{-\frac{1}{d}}\Big), d> 4.
\end{aligned}
\right.
\label{t_def}
\end{equation}
\end{theorem}

The proof of Theorem~\ref{theorem_2} is the same as the proof of \cite[Theorem 2.2]{xia2025generalized} except for the convergence rate function $h(N, d)$ when $d>4$, which we shall omit here for simplicity. Theorem~\ref{theorem_2} greatly improves the convergence rate estimation in \cite[Theorem 2.2]{xia2025generalized} and \cite[Theorem 4.3]{xia2024local} when the random field model is multidimensional but noise is heterogeneous across different dimensions. 

To better illustrate the factor $\E[h(N(\bm{x}, \delta), d)]$ characterizing the convergence rate w.r.t. the number of samples in Eq.~\eqref{theorem2_result}, consider the case when $\bm{x}$ is uniformly distributed in the unit cube $(-1, 1]^n$ and $d>4$. Then, $N(\bm{x}, \delta)\approx N\delta^n$, where $N$ is the total number of training data and 
\begin{equation}
\begin{aligned}
        h(N(\bm{x}, \delta), d)&\approx2\Big((\delta^nN)^{-\frac{1}{4}} + \big(\prod_{i=1}^d\big(\frac{\sigma_i}{\sigma_1}) \delta^nN\big)^{-\frac{1}{d}}\Big)
\end{aligned}
    \label{h_expression}
\end{equation}
Specifically, we assume that noise across different directions is heterogeneous and Eq.~\eqref{hetero_noise} holds. Using Eq.~\eqref{d_maximum}, we can further simplify Eq.~\eqref{h_expression} to:
\begin{equation}
\begin{aligned}
        h(N(\bm{x}, \delta), d)&\lesssim 2\delta^{-\frac{n}{4}} N^{-\frac{1}{4}}+2\exp\Big(-\frac{c_0d}{2}\Big)N^{-\frac{1}{d}}\delta^{-\frac{n}{d}}\\
        &\leq 2\Big(\delta^{-\frac{n}{4}} N^{-\frac{1}{4}} + \delta^{-\frac{n}{d}}N^{-\sqrt{\frac{c_0}{2\log N}}}\exp\big(-\sqrt{\frac{\log Nc_0}{2}}\big)\Big).
\end{aligned}
\label{h_simplified}
\end{equation}
Without loss of generality, we assume that the neighborhood size $\delta<1$ in Eq.~\eqref{theorem2_result}.
Plugging Eq.~\eqref{h_simplified} into Eq.~\eqref{theorem2_result}, we conclude that:
\begin{equation}
     \begin{aligned}
     \E\Big[\big|W_2^2(\bm{y}_{\bm{x}}, \hat{\bm{y}}_{\bm{x}}) - W_{2, \delta}^{2, \text{e}}(\bm{y}_{\bm{x}}, \hat{\bm{y}}_{\bm{x}})\big|\Big] &\lesssim \frac{4M_0}{\sqrt{N}} + 16CM_0\Big(\delta^{-\frac{n}{4}} N^{-\frac{1}{4}} \\
     &\hspace{-2cm}+ \delta^{-\frac{n}{d}}N^{-\sqrt{\frac{c_0}{2\log N}}}\exp\big(-\sqrt{\frac{\log Nc_0}{2}}\big)\Big) + 8\sqrt{M_0}L\delta\\
     &\hspace{-5cm}\leq \frac{4M_0}{\sqrt{N}} + 16CM_0\delta^{-\frac{n}{4}}\Big( N^{-\frac{1}{4}}+ N^{-\sqrt{\frac{c_0}{2\log N}}}\exp\big(-\sqrt{\frac{\log Nc_0}{2}}\big)\Big) + 8\sqrt{M_0}L\delta
     \end{aligned}
     \label{convergence_result}
\end{equation}
Specifically, on the RHS of the inequality~\eqref{convergence_result}, the factor $N^{-\frac{1}{4}}+ N^{-\sqrt{\frac{c_0}{2\log N}}}\exp\big(-\sqrt{\frac{\log Nc_0}{2}}\big)$ characterizing the rate of convergence w.r.t. the number of training samples does not depend on the dimensionality $d$ of the random field model $\bm{y}_{\bm{x}}$ in Eq.~\eqref{model_objective}.

Finally, when some components of $\bm{y}_{\bm{x}}$ are categorical, using the refined estimate of the convergence rate in Theorem~\ref{theorem1}, the following corollary for the generalized squared $W_2$ distance generalization error bound holds.
\begin{corollary}
\rm
    \label{corollary_2}
    Suppose the assumptions in Theorem~\ref{theorem_2} hold. Furthermore, we assume that in Eqs.~\eqref{model_objective} and \eqref{model_objective_approximate}, the first $d_1$ components of $\bm{y}_{\bm{x}}$ and $\hat{\bm{y}}_{\bm{x}}$ are continuous while the last $d-d_1$ components are categorical. Then, the following generalization error bound holds:
    \begin{equation}
    \E\Big[\big|\hat{W}_2^2(\bm{y}_{\bm{x}}, \hat{\bm{y}}_{\bm{x}}) - \hat{W}_{2, \delta}^{2, \text{e}}(\bm{y}_{\bm{x}}, \hat{\bm{y}}_{\bm{x}})\big|\Big] \leq \frac{4M_0}{\sqrt{N}} + 8CM_0\E\big[h(N(\bm{x}, \delta), {d})\big] + 8\sqrt{M_0}L\delta
        \label{theorem2_result1}
    \end{equation}
    where $\hat{W}_{2, \delta}^{2, \text{e}}(\bm{y}_{\bm{x}}, \hat{\bm{y}}_{\bm{x}})$ is the generalized local squared $W_2$ loss function defined in \cite[Eq.(2.20)]{xia2025generalized}, and $\hat{W}_{2}^{2}(\bm{y}_{\bm{x}}, \hat{\bm{y}}_{\bm{x}})$ is the squared generalized $W_2$ distance between the two random fields $\bm{y}_{\bm{x}}$ and $\hat{\bm{y}}_{\bm{x}}$ defined in \cite[Eq.~(2.18)]{xia2025generalized}.
    $M_0$ is the upper bound for $\bm{y}_{\bm{x}}$ and $\hat{\bm{y}}_{\bm{x}}$ in Eq.~\eqref{upper_boundM}, $C$ is a constant, $N$ is the total number of data points $\{(\bm{x}, \bm{y}_{\bm{x}})\}$, and $L$ is the Lipschitz constant in Eq.~\eqref{l_condition}. In Eq.~\eqref{theorem2_result1}, $h(N, d)$ is the same as that defined in Eq.~\eqref{t_def}.
\end{corollary}
The proof of Corollary~\ref{corollary_2} is the same as the proof of \cite[Theorem 2.2]{xia2025generalized}, and the only difference lies in the form of the convergence rate function $h(N,d)$ in Eq.~\eqref{t_def}.

\section{Numerical experiments}
\label{numerical_experiments}
\begin{figure}
    \centering
\includegraphics[width=0.9\linewidth]{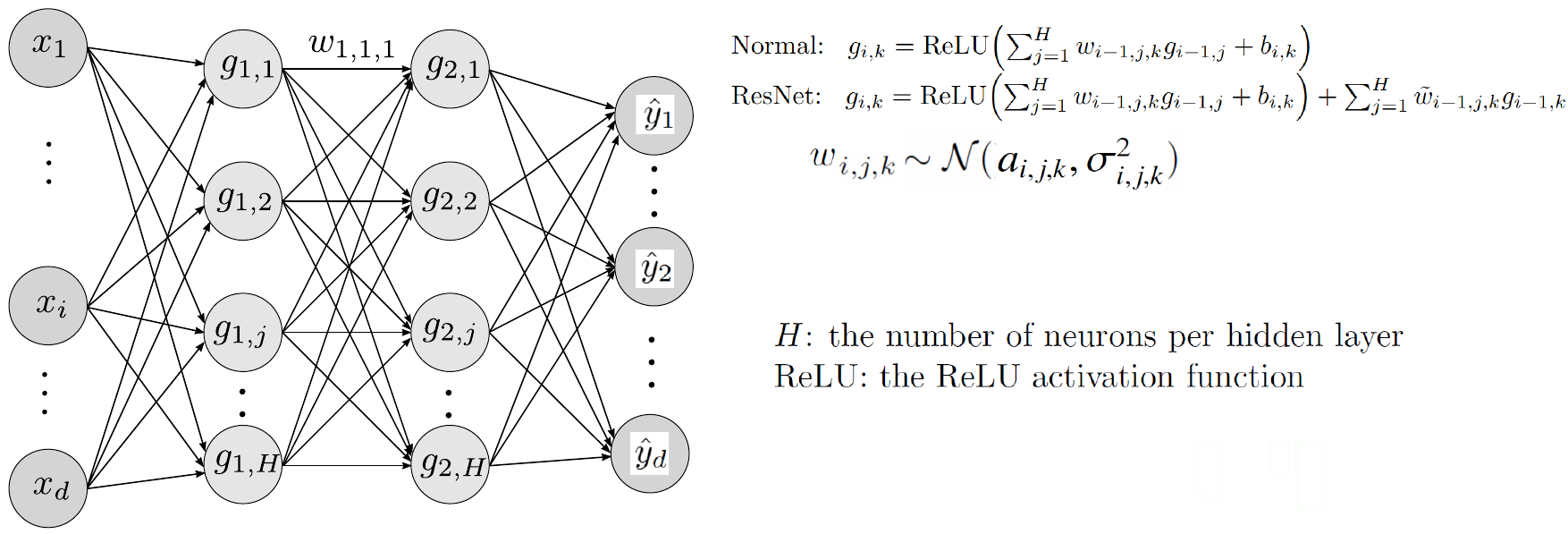}
    \caption{The structure of the SNN model used in this work. In the SNN model, for each input $x$, the weights $w_{i, j, k}\sim\mathcal{N}(a_{i, j, k}, \sigma_{i, j, k}^2)$ are independently sampled. ReLU means the \texttt{ReLU} activation function and may be replaced with other activation functions. Such an SNN, with appropriate numbers of widths and heights, has proven to be capable of approximating the random field model Eq.~\eqref{model_objective} up to any accuracy in the $W_2$ metric \cite{xia2025generalized, xia2025new}. 
    Either the normal feedforward structure or the ResNet \cite{he2016deep} structure is used for forward propagation. When using the ResNet structure, the additional weights $\tilde{w}_{i, j, k}$ are deterministic. }
    \label{fig:snn}
\end{figure}

In this section, we conduct numerical experiments to test our proposed generalized $W_2$ method. To boost efficiency, given $N$ observed data $\{(\bm{x}_i, \bm{y}_{i})\}_{i=1}^N$, instead of using the original local squared $W_2$ loss function \cite[Eq. 3]{xia2024local}, we adopt a minibatch technique and use the following loss function:
\begin{equation}
          \frac{1}{|X_0|}\sum_{\bm{x}\in X_0} W_2^2(\mu_{\bm{x}, \delta}^{\text{e}}, \hat{\mu}_{\bm{x}, \delta}^{\text{e}}),
          \label{updated_loss}
\end{equation}
where are the empirical probability measures of $\bm{y}(\tilde{\bm{x}};\omega)$ and $\hat{\bm{y}}(\tilde{\bm{x}};\hat{\omega})$ conditioned on $\|\tilde{\bm{x}}-\bm{x}\|_2\leq\delta$, and $X_0\subseteq X\coloneqq\{\bm{x}_i\}_{i=1}^N$ ($|X_0|$ refers to the number of its elements) is randomly chosen such that for each 
\begin{equation}
    \bm{x}\in X_0, N(\bm{x},\delta)\geq N_0.
    \label{n0_def}
\end{equation}
Therefore, we do not use samples whose neighborhoods contain too few samples when training the SNN (in all examples, we set $N_0=4$ in Eq.~\eqref{n0_def}).
$X_0$ is renewed and randomly selected again after every fixed number of training epochs. Numerical experiments in all examples are conducted using Python 3.11 on a desktop with a 32-core Intel®
i9-13900KF CPU. The structure of the SNN we use is similar to that used in \cite{xia2025generalized} and is described in Fig.~\ref{fig:snn}. SNNs in Fig.~\ref{fig:snn} have the capability to approximate the unknown random field model Eq.~\eqref{model_objective} up to any accuracy under the $W_2$ metric (shown in \cite{xia2025new, xia2025generalized}). Additionally, we can show that the SNN in Fig.~\ref{fig:snn} is also robust against perturbations in the biases $\{b_{i, k}\}$ as well as the means $\{a_{i, j, k}\}$ and variances $\sigma_{i, j, k}$ of the weights $w_{i-1, j, k}$, making them a good surrogate model for approximating the random field model Eq.~\eqref{model_objective}. The proof is given in Appendix~\ref{snn_robust}.

A pseudocode of our generalized $W_2$ approach to train the SNN in Fig.~\ref{fig:snn} by minimizing the loss function Eq.~\eqref{updated_loss} is given in Algorithm~\ref{algorithm_1}, which is similar to \cite[Algorithm 1]{xia2025generalized}. Training hyperparameters and settings for each example are listed in Table~\ref{tab:setting} in Appendix~\ref{training_details}. 
\begin{algorithm}
\footnotesize
\caption{\footnotesize The pseudocode of the local squared $W_2$ approach to train an SNN.}
\begin{algorithmic}
  \STATE Given $N$ observed data $\{(\bm{x}_i, \bm{y}_i), i=1,..., N\}$, the size of the neighborhood $\delta$, the size of a minibatch $n$, the number of epochs for updating a minibatch $\text{epoch}_{\text{update}}$ (set to 20 for all examples),  and the maximal epochs $\text{epoch}_{\max}$.
    \STATE Initialize the SNN in Fig.~\ref{fig:snn}.
\STATE For each $\bm{x}_i$, find samples in its neighborhood $B_i\coloneqq \{\bm{x}_j: \|\bm{x}_j-\bm{x}_i\|_2\leq \delta\}$.
    \STATE Input $\{\bm{x}_i\}, i=1,...,N$ into the neural network model to obtain predictions $\{\hat{\bm{y}}_i\}, i=1,...,N$.
  \FOR{$j=0,1,...,\text{epoch}_{\max}-1$}
  \IF{$j~\%~\text{epoch}_{\text{update}}$ == 0}
  \STATE Randomly choose $n$ samples from $\{(\bm{x}_i, \bm{y}_i), i=1,..., N, N(\bm{x}_i, \delta)\geq N_0\}$ to get a new $X_0$ in Eq.~\eqref{updated_loss}.
  \ENDIF
  \STATE Calculate the loss function Eq.~\eqref{updated_loss}.
  \STATE Perform gradient descent to minimize the loss function
     and update the parameters in the SNN.
\STATE Resample the weights in the SNN using the updated means and variances of weights.
    \STATE Input $\{\bm{x}_i\}, i=1,...,N$ into the updated SNN to obtain predictions $\{\hat{\bm{y}}_i\}, i=1,...,N$.  (for each $\bm{x}_i$, the weights in the SNN are sampled independently).
    \ENDFOR
  \RETURN The trained SNN
\end{algorithmic}
\label{algorithm_1}
\end{algorithm}

\begin{example}
\label{example1}
\rm
In this example, we work on a synthetic data set to investigate how the dimensionality of the uncertainty model $\bm{y}$ in Eq.~\eqref{model_objective} and the dimensionality of the noise affect the reconstruction of it. Here, we generate a synthetic data set $\{(\bm{x}_i, \bm{z}(\bm{x}_i), \bm{c})\}_{i=1}^N$ using the \texttt{make\_regression} function in the \texttt{sklearn.datasets} package of Python as the training data. Here, $\bm{x}_i=(x_{1, i}, x_{2, i})\sim\mathcal{N}(0, \mathbb{I}_2)$, where $\mathbb{I}_2\in\mathbb{R}^{2\times 2}$ is the identity matrix, and $\bm{z}(\bm{x}_i)\coloneqq (z_j^k(\bm{x}_i))_{j=1,...d, k=1,...,d_0+1}\in\mathbb{R}^{d\times (d_0+1)}$ with each component in $\bm{z}(\bm{x})$ being a linear function on $x_{1, i}$ and $x_{2, i}$. $\bm{c}\coloneqq (c)_{j,k,\ell}\in\mathbb{R}^{d\times (\ell+1)\times2}$ records the randomly sampled coefficients when using $(x_{1, i}, x_{2, i})$ to calculate $\bm{z}(\bm{x}_i)$:
\begin{equation}
    z_{j}^k(\bm{x}_i) = c_{j,k,1}x_{i, 1} + c_{j,k,2}x_{i, 2}.
    \label{linear}
\end{equation}
We set:
\begin{equation}
\bm{y}_{\bm{x}_i}\coloneqq (y_{\bm{x}_i, 1},...,y_{\bm{x}_i, d}),\,\,
y_{\bm{x}_i, j}\coloneqq \sin\Big(\frac{1}{16}\big(z_j^1(\bm{x}_i) + \sum_{k=2}^{{d_0+1}}{z}_j^k(\bm{x}_i)\epsilon_j\big)\Big), \,\,j=1,...,d,
\label{example1_model}
\end{equation}
where $\epsilon_j\sim\mathcal{N}(0, \sigma_j^2)$ are independent random variables. In Eq.~\eqref{example1_model}, $z_j^k(\bm{x}_i)$ denotes the element in the $j^{\text{th}}$ row and $k^{\text{th}}$ column of the matrix $\bm{z}(\bm{x}_i)$.
For every $\bm{x}$, all possible values of $\bm{y}$ lie in a $d_0$-dimensional manifold in $\mathbb{R}^d$. 
We carry out the following three experiments:
\begin{enumerate}
\item We set $d_0=3$ and vary $d=3,...,23$ to investigate how the dimensionality of $\bm{y}$ in Eq.~\eqref{example1_model} affects the reconstruction of it. $\sigma_j\equiv0.1$ for all different values of $j$.
\item We set $d_0=d=3,...,12$ and $\sigma_j=0.1\exp(-j)$ to explore how heterogeneous noise across different dimensions affects the reconstruction of Eq.~\eqref{example1_model}. Specifically, we compare minimizing our improved loss function Eq.~\eqref{updated_loss} versus minimizing the previous local squared $W_2$ loss function \cite[Eq. (3.1)]{xia2025generalized} for training SNNs.
    \item We set $d=12$ and vary the value of $d_0=3,...,10$ to investigate how the dimensionality of noise affects the reconstruction of Eq.~\eqref{example1_model}. We consider two subcases: i) using a fixed $\sigma_j=0.1$ for all $j$ and ii) setting $\sigma_j=0.1\sqrt{\frac{3}{d}}$ for all $j$ so that the total noise intensity summed over all dimensions is expected to be the same.
\end{enumerate}
As the testing set, we randomly sample 100 $\{(x_1^i, x_2^i)\}_{i=1}^{100}, x_1^i, x_2^i\sim\mathcal{U}(-\frac{1}{4}, \frac{1}{4})$. At each $(x_1^i, x_2^i)$, we independently generate 20 $\bm{y}_{\bm{x}_i}$ using Eqs.~\eqref{linear} and \eqref{example1_model}.
To evaluate the accuracy of the reconstructed $\hat{\bm{y}}_{\bm{x}}\coloneqq(\hat{y}_{\bm{x}, 1},...,\hat{y}_{\bm{x}, d})$ on the testing set, we use the following two metrics:
\begin{equation}
\begin{aligned}
    &\text{Average relative error in mean}\coloneqq \frac{1}{d|X_t|}\sum_{i=1}^d\sum_{\bm{x}\in X_t}\frac{|\E[\hat{y}_{\bm{x}, i}]- \E[y_{\bm{x}, i}]|}{|\E[y_{\bm{x}, i}]|},\\
    &\hspace{-0.26in}\text{Average relative error in standard deviation (SD)}\coloneqq \frac{1}{d|X_t|}\sum_{i=1}^d\sum_{\bm{x}\in X_t}\frac{|\text{SD}[\hat{y}_{\bm{x}, i}]- \text{SD}[y_{\bm{x}, i}]|}{|\text{SD}[y_{\bm{x}, i}]|},
\end{aligned}
\label{relative_error}
\end{equation}
where $X_t$ refers to the testing set ($|X_t|$ is the number of elements in it). In \cite{xia2024local, xia2025generalized}, it has been shown that directly minimizing the local squared $W_2$ loss function, such as Eq.~\eqref{updated_loss}, to train SNNs can better reconstruct the random field model Eq.~\eqref{model_objective} than other approaches commonly used in UQ, such as the Bayesian neural networks, the Wasserstein GAN approach, and the evidential learning method. Here, we shall benchmark the local squared $W_2$ approach for training SNNs against two additional generative UQ methods, the conditional variational autoencoder approach \cite{Kingma2013, lopez2017conditional} and the conditional normalization flow approach \cite{papamakarios2017masked, winkler2019learning}, for reconstructing the model Eq.~\eqref{example1_model}. When implementing these benchmark methods, the number of hidden layers is two since both the conditional variational autoencoder approach and the conditional normalization flow method need two neural networks. Other hyperparameters in the neural network and the optimization hyperparameters are the same as those used for minimizing the loss Eq.~\eqref{updated_loss} to train the SNN (listed in Table~\ref{tab:setting} under Example 1).



    \begin{figure}[H]
    \centering
\includegraphics[width=\linewidth]{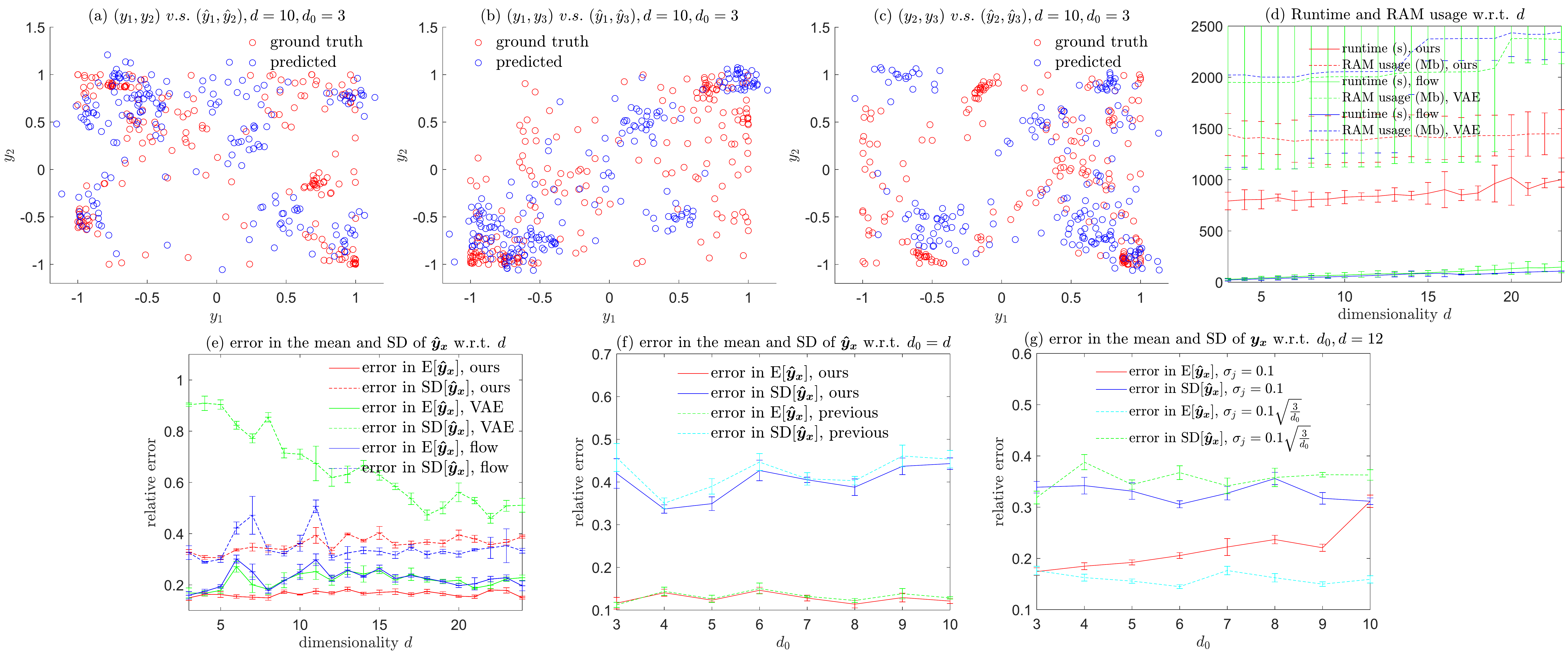}
    \caption{(a)(b)(c) the joint distributions of the ground truth $(y_{\bm{x}, 1}, y_{\bm{x}, 2})$, $(y_{\bm{x}, 1}, y_{\bm{x}, 3})$, and $(y_{\bm{x}, 2}, y_{\bm{x}, 3})$ versus the predicted $(\hat y_{\bm{x}, 1},  \hat y_{\bm{x}, 2})$, $(\hat y_{\bm{x}, 1},\hat y_{\bm{x}, 3})$, and $(\hat y_{\bm{x}, 2}, \hat y_{\bm{x}, 3})$ for 10 different $\bm{x}$ on the testing set (at each $\bm{x}$, there are 20 independently generated $\bm{y}_{\bm{x}}$ and $\hat{\bm{y}}_{\bm{x}}$, respectively). In (a), (b), and (c), $\sigma_j\equiv 0.1$. (d) the runtime and RAM usage w.r.t. the dimensionality $d$ of the random field model Eq.~\eqref{example1_model} (case 1). ``ours" refers to directly minimizing the local squared $W_2$ loss function Eq.~\eqref{updated_loss} to train the SNN in Fig.~\ref{fig:snn}. ``VAE" denotes the conditional variational encoder approach, while ``flow" denotes the conditional normalization flow method. (e) errors in the mean and SD of the predicted $\hat{\bm{y}}_{\bm{x}}$ w.r.t. the dimensionality $d$ of the random field model Eq.~\eqref{example1_model} (case 2). (f) errors in the mean and SD of the predicted $\hat{\bm{y}}_{\bm{x}}$ w.r.t. the dimensionality $d_0=d$ of the random field model Eq.~\eqref{example1_model} (case 3). (g) errors in the mean and SD of the predicted $\hat{\bm{y}}_{\bm{x}}$ w.r.t. the dimensionality $d_0$ of noise Eq.~\eqref{example1_model} (case 3).}
    \label{fig:example1}
\end{figure}

In Fig.~\ref{fig:example1} (a)(b)(c), we plot the ground truth joint distributions of $(y_{\bm{x}, 1}, y_{\bm{x}, 2})$, $(y_{\bm{x}, 1}, y_{\bm{x}, 3})$, and $(y_{\bm{x}, 2}, y_{\bm{x}, 3})$ versus the joint distributions of the predicted $(y_{\bm{x}, 1}, y_{\bm{x}, 2})$, $(y_{\bm{x}, 1}, y_{\bm{x}, 3})$, and $(y_{\bm{x}, 2}, y_{\bm{x}, 3})$ when $d=10, d_0=3, \sigma_j\equiv 0.1$ for 10 different $\bm{x}$ in the testing set. The ground truth joint distributions can be matched well by the predicted distributions.
As the dimensionality $d$ of the random field model Eq.~\eqref{example1_model} increases, the RAM usage seems unaffected while the runtime slightly increases (shown in Fig.~\ref{fig:example1} (d)), implying that the dimensionality of the random field model will not lead to a significant increase in computational costs of the local squared $W_2$ approach. Compared to the conditional variational encoder and the conditional normalization flow approaches, the RAM usage of our method is smaller, but the runtime is longer, resulting from the $O(N^3)$ computational complexity of evaluating the Wasserstein-distance between two empirical probability measures consisting of $N$ samples. 
In Fig.~\ref{fig:example1} (e), the errors in the mean and SD
of the predicted $\hat{\bm{y}}_{\bm{x}}$ do not increase much as the dimensionality $d$ of the random field model Eq.~\eqref{example1_model} increases as long as $d_0$, the dimensionality of noise, does not change with $d$. 
Furthermore, when $d_0=d$ but noise is strongly heterogeneous across different dimensions (case 2),  the errors in the mean and SD  
of the predicted $\hat{\bm{y}}_{\bm{x}}$ do not change much as the dimensionality $d$, either. These results indicate that SNNs can be trained to accurately reconstruct multidimensional random field models, provided the noise lies on a low-dimensional manifold or is strongly heterogeneous across different dimensions. Compared to previous loss functions \cite[Eq. (3.1)]{xia2025generalized}, minimizing our improved loss function Eq.~\eqref{updated_loss}, which ignores training samples whose neighborhoods contain too few other training samples, leads to improved accuracy of the predicted SD of $\hat{\bm{y}}_{\bm{x}}$. Additionally, the conditional variational encoder approach yields an inaccurate reconstructed SD and an inaccurate predicted mean, while the error in the predicted mean from the conditional normalization flow method is much larger than the predicted mean of our approach. 
Finally, as shown in Fig.~\ref{fig:example1} (g), when $d=12$, the error in the mean of the predicted $\bm{y}_{\bm{x}}$ increases w.r.t. the dimensionality of noise $d_0$ if $\sigma_j\equiv0.1$ is a constant. Therefore, as the dimensionality of noise increases, it will be harder to reconstruct the distribution of $\bm{y}_{\bm{x}}$ in Eq.~\eqref{example1_model} if the strength of noise is constant in each dimension.
However, when $\sigma_j=0.1\sqrt{\frac{3}{d_0}}$ such that noise in each dimension decreases with $d_0$ and $\sum_{j=1}^{d_0}\sigma_j^2\equiv 0.03$, the relative error in the mean and SD of $\hat{\bm{y}}_{\bm{x}}$ does not change much w.r.t. $d_0$.

\end{example}

Next, we consider reconstructing a high-dimensional ODE system with uncertain parameters.
\begin{example}
\label{example3_case}
\rm

We consider the following 96-dimensional ODE system in \cite[Example 4.5]{sonday2011eigenvalues} describing the dynamics of a system of damped linear oscillators with state $(x, v)\in \mathbb{R}^{48} \times \mathbb{R}^{48}$ governed by the system of ODEs:
\begin{equation}
\begin{aligned}
    \frac{\d x_j}{\d t} &= v_j, \\
    \frac{\d v_j}{\d t} &= \frac{1}{50}(x_{j-1} - x_j) + \frac{1}{50}(x_{j+1} - x_j) - c_jv_j, j=1,...,48
    \end{aligned}
    \label{example3}
\end{equation}
with $x_0=x_{49}=v_0=v_{49}\equiv0$.   The initial condition $(\hat{\bm{x}}(0),\hat{\bm{v}}(0))^T\sim\mathcal{N}(\bm{1},\sigma_0^2\mathbb{I}_{96})$, where $\mathbb{I}_{96}$ refers to the identity matrix in $\mathbb{R}^{96\times 96}$. Different from \cite[Example 4.5]{sonday2011eigenvalues} in which $c_j\equiv c\sim \exp(0.25\xi-1.6)$ with $\xi\sim\mathcal{N}(0, 1)$, we use heterogeneous $c_j$ for different $v_j$:
\begin{equation}
    c_j=  \exp\big(\tfrac{\xi_j}{4}-1.6\big), j=1,...,d-1,\,\, c_j=\exp\big(\tfrac{\xi_d}{4}-1.6\big), j\geq d,
    \label{example3_sensitivity}
\end{equation}
where each $\xi_j$ is independently sampled from $\mathcal{N}(0, \sigma^2)$. We use the following approximate ODE to reconstruct Eq.~\eqref{example3}:
\begin{equation}
    \frac{\d\begin{pmatrix}
    \hat{\bm{x}} \\
    \hat{\bm{v}}
\end{pmatrix}}{\d t} = \hat{\bm{f}}(\hat{\bm{x}},\hat{\bm{v}};\hat{\theta}), \,\hat{\bm{x}}, \hat{\bm{v}}\in\mathbb{R}^{48},
\label{reconstructed_dyn}
\end{equation}
where $\hat{\bm{f}}$ refers to the SNN that approximates Eq.~\eqref{example3} and is trained through minimizing a time-averaged version of the loss function Eq.~\eqref{updated_loss}:
\begin{equation}
    \frac{1}{N_T}\frac{1}{|X_0|}\sum_{i=1}^{N_T}\sum_{\bm{x}\in X_0} W_2^2(\mu_{\bm{x}, \delta}^{\text{e}}(t_i), \hat{\mu}_{\bm{x}, \delta}^{\text{e}}(t_i)).
    \label{time_decoupled}
\end{equation}
The uncertainty parameters $\hat{\theta}$ in the SNN will only be sampled once at $t=0$ when generating one trajectory of $(\hat{\bm{x}}, \hat{\bm{v}})^T$. For each trajectory of $\hat{\bm{y}}$, $\hat{\theta}$ will be sampled independently.
In Eq.~\eqref{time_decoupled}, $\mu_{\bm{x}, \delta}^{\text{e}}(t_i)$ and $\hat{\mu}_{\bm{x}, \delta}^{\text{e}}(t_i)$ refer to the probability measures associated with $\bm{y}(t_i)\coloneqq(\bm{x}(t_i),\bm{v}(t_i))^T$ and $\hat{\bm{y}}(t_i)\coloneqq(\hat{\bm{x}}(t_i),\hat{\bm{v}}(t_i))^T$, respectively.
We carry out numerical experiments to explore:
\begin{enumerate}
    \item How the dimensionality $d$ of the uncertain latent parameters affects the reconstruction of Eq.~\eqref{example3}. In this case, $\sigma=1$ and $\sigma_0=0.01$ are constants.
    \item How the strength of noise $\sigma$ for each $\xi_j$ in Eq.~\eqref{example3_sensitivity} affects the reconstruction of Eq.~\eqref{example3}. $d=5$ and $\sigma_0=0.01$ in this scenario.
    \item How the uncertainty in the initial condition $\sigma_0$ affects the reconstruction of Eq.~\eqref{example3}. In this case, $d=5$ and $\sigma=1$.
\end{enumerate}
In all cases, we set $t_i=i\Delta t$ and $N_T = 30$ in Eq.~\eqref{time_decoupled}.
To evaluate the errors in the reconstructed trajectories $(\hat{\bm{x}}(t), \hat{\bm{v}}(t))^T$ as well as in the learned dynamics $\hat{\bm{f}}$ in Eq.~\eqref{reconstructed_dyn}, we use the following error metrics: 
\begin{equation}
    \text{error in } \bm{\hat{y}}\coloneqq \frac{\E\Big[W_2^2\big(\mu_{ t}, \hat{\mu}_{ t}\big)\Big]}{\E\big[\|\bm{y}(t)\|^2\big]},\,\, \text{error in } \hat{\bm{f}}\coloneqq \frac{\E\Big[W_2^2\big(\eta_{\bm{y}(t),t}, \hat{\eta}_{\bm{y}(t), t}\big)\Big]}{\E\big[\|\bm{f}(\bm{y}(t); \theta)\|^2\big]},
\end{equation}
where $\mu_t$ and $\hat{\mu}_t$ are probability measures of $\bm{y}(t)=(\bm{x}(t),\bm{v}(t))^T$ and $(\hat{\bm{x}}(t), \hat{\bm{v}}(t))^T$, respectively. $\bm{f}(\bm{y}(t);\theta)\coloneqq \frac{\d\bm{y}(t)}{\d t}$ and $\theta$ denotes the set of uncertain parameters $c_j$ in Eq.~\eqref{example3_sensitivity}.  $\eta_{\bm{y}(t), t}$ and $\hat{\eta}_{\bm{y}(t), t}$ denote the probability measures of $\bm{f}(\bm{y}(t);\theta)$ and $\hat{\bm{f}}(\bm{y}(t);\hat{\theta})$ in Eq.~\eqref{reconstructed_dyn}, respectively.

    \begin{figure}[H]
    \centering
\includegraphics[width=\linewidth]{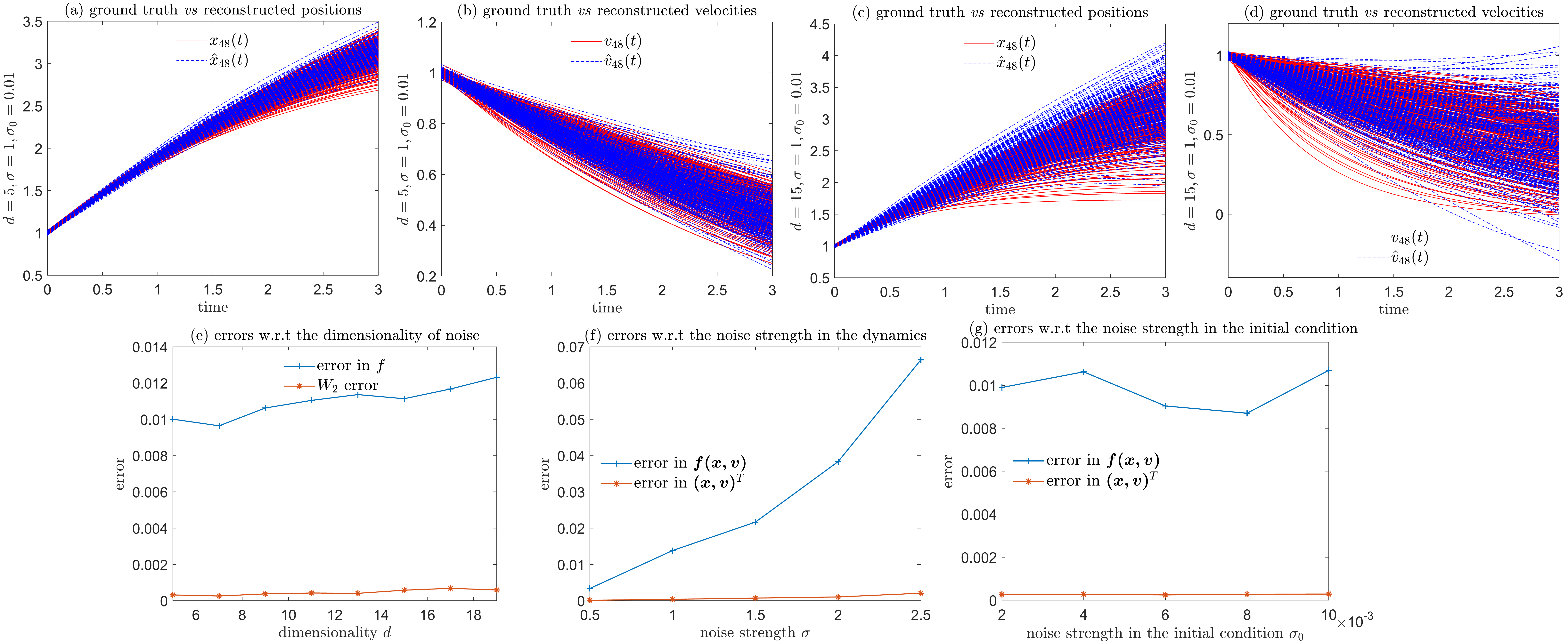}
    \caption{(a)(b) the trajectories of the reconstructed position and velocity $\hat x_{48}(t)$ and $\hat v_{48}(t)$ versus the ground truth position and velocity $ x_{48}(t)$ and $ v_{48}(t)$ when $d=5, \sigma=1, \sigma_0=0.01$. (c)(d) the trajectories of the reconstructed position and velocity $\hat x_{48}(t)$ and $\hat v_{48}(t)$ versus the ground truth position and velocity $ x_{48}(t)$ and $ v_{48}(t)$ when $d=5, \sigma=2.5, \sigma_0=0.01$. (e) errors in $(\hat{\bm{x}}(t), \hat{\bm{v}}(t))^T$ and in $\hat{\bm{f}}$ w.r.t. the dimensionality of uncertain parameters $d$. (f) errors in $(\hat{\bm{x}}(t), \hat{\bm{v}}(t))^T$ and in $\hat{\bm{f}}$ w.r.t. the strength of noise in the dynamics $\sigma$. (g) errors in $(\hat{\bm{x}}(t), \hat{\bm{v}}(t))^T$ and in $\hat{\bm{f}}$ w.r.t. the noise in the initial condition $\sigma_0$.}
    \label{fig:example3}
\end{figure}

From Fig.~\ref{fig:example3} (a)(b)(c)(d), when the strength of noise is small, the reconstructed trajectories align well with ground truth trajectories, and when the strength of noise gets larger, it is more difficult for the reconstructed trajectories to align well with ground truth trajectories. As shown in Fig.~\ref{fig:example3} (e), as the dimensionality $d$ of uncertainty parameters increases, errors in both the reconstructed trajectories $(\hat{\bm{x}}, \hat{\bm{v}})^T$ and in the learned dynamics $\hat{\bm{f}}$ increase. This indicates that when reconstructing noisy dynamical systems with uncertain latent parameters, an increased dimensionality of those uncertain parameters leads to a less accurate reconstruction of the underlying dynamics. In Fig.~\ref{fig:example3} (f), the error in the learned dynamics $\hat{\bm{f}}$ increases as the strength of noise in the noisy dynamical systems increases, implying that it is more difficult to reconstruct a noisy dynamical system when noise in such a system gets stronger. Finally, training the SNN through minimizing the loss function Eq.~\eqref{time_decoupled} to learn $\hat{\bm{f}}$ in Eq.~\eqref{reconstructed_dyn} is insensitive to the strength of noise in the initial condition, as shown in Fig.~\ref{fig:example3} (g). 
\end{example}

\section{Summary and conclusion}
\label{summary}
In this work, we performed an analysis on the scalability of training SNNs to reconstruct multidimensional random field models by minimizing Wasserstein-distance-type loss functions. Specifically, we proved that the convergence rate of the empirical probability measure to the ground truth probability measure might not explicitly depend on the dimensionality of the model when noise is heterogeneous across different dimensions, relieving the ``curse of dimensionality" for learning multidimensional random fields from a finite number of data. We improved the local squared $W_2$ method for training SNNs to make the trained SNN more robust in scenarios when data is sparse. We also proved that the distribution of the output of the SNN model is robust against perturbations in its biases as well as means \& variances of their weights, making it a good surrogate model for reconstructing random field models. Through numerical experiments, we showed that when noise was heterogeneous across different dimensions or lay in a low-dimensional manifold, the local squared $W_2$ method could train an SNN that approximated the unknown multidimensional random field model well. Furthermore, the local squared $W_2$ method outperformed several benchmark generative approaches in UQ for reconstructing multidimensional random field models. 

In the future, it is a prospective research field to take into account how to incorporate prior knowledge of noise when training the SNN, \text{e.g.}, enforcing some physics-informed constraints in the loss function \cite{karniadakis2021physics}. As another direction, it is worthwhile considering the application of the entropic regularized Wasserstein distance and applying the Sinkhorn algorithm \cite{cuturi2013sinkhorn} to boost efficiency and reduce computational costs. Specifically, a discussion on the $W_2$ loss landscape w.r.t. parameters in the SNN is helpful.
Additionally, it is worth further exploration on optimizing the structure of the SNN in Fig.~\ref{fig:snn} to boost efficiency.
Finally, it is promising to test our Wasserstein-distance-based SNN training approach on more high-dimensional UQ tasks to learn random fields or models with uncertainty.

\section*{Data Availability}
 All code and data will be made publicly available upon acceptance of this manuscript.

\appendix
\section{Proof for Theorem~\ref{theorem1}}
\label{proof_thm1}

The proof of Theorem~\ref{theorem1} builds upon the proof of \cite[Theorem 1]{fournier2015rate}. Without loss of generality, we assume that $\sigma_1=1$.
 Similar to \cite[Lemma 5]{fournier2015rate}, we introduce the following notations.
\begin{enumerate}
\item[(a)] For $\ell \geq 0$, we denote by $\mathcal{P}_\ell$ the smallest set that consists of non-overlapping translations of $(-2^{-\ell}, 2^{-\ell}]^d$ such that $S_d\coloneqq\otimes_{i=1}^d(-\sigma_i, \sigma_i]\subseteq\cup_{C\in\mathcal{P}_{\ell}} C$. 
For two probability measures $\mu, \nu$ on $S_d$, we introduce
\[
D_2(\mu, \nu) := \frac{2^2 - 1}{2} \sum_{\ell\geq 1} 2^{-2\ell} \sum_{F\in\mathcal{P}_\ell} |\mu(F) - \nu(F)|,
\]
which defines a distance on $\mathcal{P}(S_d)$, always bounded by 1.

\item[(b)] We introduce $B_0 := S_d$ and 
\begin{equation}
    B_n := \otimes_{i=1}^d(-2^n\sigma_i, 2^n\sigma_i] \setminus \otimes_{i=1}^d(-2^{n-1}\sigma_i, 2^{n-1}\sigma_i], n\geq 1.
    \label{Bn_def}
\end{equation} 
For $\mu \in \mathcal{P}(\mathbb{R}^d)$ and $n \geq 0$, we denote by $R_{B_n}\mu$ the probability measure on $S_d$ defined as the image of $\mu|_{B_n}/\mu(B_n)$ by the map $x \mapsto x/2^n$. For two probability measures $\mu, \nu$ on $\mathbb{R}^d$ and for $p > 0$, we introduce
\begin{equation}
\mathcal{D}_2(\mu, \nu) := \sum_{n\geq 0} 2^{2n} \big[|\mu(B_n) - \nu(B_n)| + (\mu(B_n) \wedge \nu(B_n))D_2(R_{B_n}\mu, R_{B_n} \nu)\big].
\label{D_def}
\end{equation}
Since $D_2(\mu,\nu) \leq 1$ on $\mathcal{P}(S_d)$, $\mathcal{D}_2$ defines a distance on $\mathcal{P}(\mathbb{R}^d)$.
\end{enumerate}

 We first assume that $\mu$ and $\nu$ are supported in $S_d$. We regard $\mu, \nu$ as two probability measures in $(-1, 1]^d$ such that $\mu((-1, 1]^d-S_d) = \nu\big((-1, 1]^d-S_d\big)=0$. 
    We infer from [16, Lemma 2] that, since the diameter of $S_d$ is less than $2\sqrt{d}$:
\begin{equation}
W_2^2(\mu, \nu) \leq \frac{\left(2\sqrt{d}\right)^2}{2} \sum_{\ell=0}^{\infty} 2^{-2\ell} \sum_{F\in\mathcal{P}_\ell} \mu(F) \sum_{\substack{C \text{ child of } F}} \left|\frac{\mu(C)}{\mu(F)} - \frac{\nu(C)}{\nu(F)}\right|
\label{t2_def}
\end{equation}
with the convention $\frac{0}{0}=0$. ``$C$ child of $F$'' means that $C \in \mathcal{P}_{\ell+1}$ and $C \subset F$.  Specifically, when $F\cap S_d=\emptyset$, $\mu(F)=\nu(F)=0$. Therefore, we have:

\begin{equation}
\begin{aligned}
W_2^2(\mu, \nu) &\leq 2d \sum_{\ell=0}^{\infty} 2^{-2\ell} \sum_{F\in\mathcal{P}_\ell} \sum_{\substack{C \text{ child of } F}} \Big[\frac{\nu(C)}{\nu(F)} |\mu(F) - \nu(F)| + |\mu(C) - \nu(C)|\Big] \\
&\leq 2d \sum_{\ell=0}^{\infty} 2^{-2\ell} \Big[ \sum_{F\in\mathcal{P}_\ell} |\mu(F) - \nu(F)| + \sum_{C\in\mathcal{P}_{\ell+1}} |\mu(C) - \nu(C)| \Big] \\
&\leq 2d\cdot 5 \sum_{\ell=1}^{\infty} 2^{-2\ell} \sum_{F\in\mathcal{P}_\ell} |\mu(F) - \nu(F)|,
\end{aligned}
\end{equation}

We next consider the case when $n\geq 1$ for $B_n$ in Eq.~\eqref{Bn_def}. For any $\epsilon>0$, there exists a coupling between $R_{B_n}\mu$ and $R_{B_n}\nu$ denoted as $\pi_n(\d x, \d y)$ for such that 
\begin{equation}
    \E_{(\bm{x}, \bm{y})\sim \pi_n}\Big[\|\bm{x}-\bm{y}\|^2\Big]< W_2^2(R_{B_n}\mu, R_{B_n}\nu)+\epsilon.
\end{equation}
Furthermore, the marginal distributions of $\pi_n(\d x, \d y)$ coincide with $R_{B_n}\mu$ and $R_{B_n}\nu$, respectively. We define $\tilde{\pi}_n(\d x, \d y)$ as the image of $\pi_n$ by the map $(x, y) \mapsto (2^n x, 2^n y)$, whose marginal distributions coincide with $\mu|_{B_n}/\mu(B_n)$ and $\nu|_{B_n}/\nu(B_n)$. Furthermore, $\tilde{\pi}_n(\d x, \d y)$ satisfies:
\begin{equation}
\hspace{-0.15in}\int_{\mathbb{R}^d\times\mathbb{R}^d} \|x - y\|^2 \tilde{\pi}_n(\d x, \d y) = 2^{2n} \int_{\mathbb{R}^d\times\mathbb{R}^d} \|x - y\|^2 \pi_n(\d x, \d y) < 2^{2n}\big(W_2^2(R_{B_n}\mu, R_{B_n}\nu)+\epsilon\big).
\end{equation}

Next, we denote $q := \frac{1}{2} \sum_{n\geq0} |\mu(B_n) - \nu(B_n)|$, and we define
\begin{equation}
    \pi(\d x, \d y) = \sum_{n\geq0} (\nu(B_n) \wedge \mu(B_n))\tilde{\pi}_n(\d x, \d y) + \frac{\alpha(\d x)\beta(\d y)}{q},
    \label{couple_def}
\end{equation}
where
\[
\alpha(\d x) := \sum_{n\geq0} (\mu(B_n) - \nu(B_n))_+ \frac{\mu|_{B_n}(\d x)}{\mu(B_n)}, \,\,\,
\beta(\d y) := \sum_{n\geq0} (\nu(B_n) - \mu(B_n))_+ \frac{\nu|_{B_n}(\d y)}{\nu(B_n)}.
\]
Using the fact that:
\[
q = \sum_{n\geq0} (\nu(B_n) - \mu(B_n))_+ = \sum_{n\geq0} (\mu(B_n) - \nu(B_n))_+
= 1 - \sum_{n\geq0} (\nu(B_n) \wedge \mu(B_n)),
\]
we can check that $\pi$ is a probability measure on $\mathcal{P}(\mathcal{R}^d\times\mathcal{R}^d)$. Furthermore, we have:
\begin{equation}
\begin{aligned}
\frac{\int_{\mathbb{R}^d\times\mathbb{R}^d} \|x - y\|^2 \alpha(\d x)\beta(\d y)}{q} &\leq \frac{1}{q} \int_{\mathbb{R}^d\times\mathbb{R}^d} 2(\|x\|^2 + \|y\|^2)\alpha(\d x)\beta(\d y) \\
&= 2 \int_{\mathbb{R}^d} \|x\|^2 \alpha(\d x) + 2 \int_{\mathbb{R}^d} \|y\|^2 \beta(\d y) \\
&\leq 2d \sum_{n\geq0} 2^{2n} \big[(\mu(B_n) - \nu(B_n))_+ + (\nu(B_n) - \mu(B_n))_+\big] \\
&= 2d \sum_{n\geq0} 2^{2n} |\mu(B_n) - \nu(B_n)|.
\end{aligned}
\end{equation}

 Since $\int_{\mathbb{R}^d\times\mathbb{R}^d} \|x - y\|^2 \tilde{\pi}_n(\d x, \d y) \leq 2^{2n} \big(W_2^2(R_{B_n}\mu, R_{B_n} \nu)+\epsilon\big)$, using the definition of $\pi$ in 
Eq.~\eqref{couple_def}, we conclude that:
 \begin{equation}
\begin{aligned}
W_2^2(\mu, \nu) &\leq \int_{\mathbb{R}^d\times\mathbb{R}^d} \|x - y\|^2 \pi(
\d x, \d y) \\
&\hspace{-0.2in}\leq 2d \sum_{n\geq0} 2^{2n} |\mu(B_n) - \nu(B_n)| + \sum_{n\geq0} 2^{2n} (\nu(B_n) \wedge \mu(B_n)) \big(W^2_2(R_{B_n}\mu, R_{B_n}\nu) +\epsilon\big)\\
&\hspace{-0.2in}\leq 2d \sum_{n\geq0} 2^{2n} \left[|\mu(B_n) - \nu(B_n)| + (\mu(B_n) \wedge \nu(B_n))\big(W^2_2(R_{B_n}\mu, R_{B_n}\nu)+\epsilon\big)\right].\\
&\hspace{-0.2in}\leq 2d \mathcal{D}_2(\mu, \nu) + 2d\sum_{n\geq0}2^{2n}(\mu(B_n) \wedge \nu(B_n))\epsilon.
\label{W2_bound}
\end{aligned}
\end{equation}
From the definition of $B_n$ in Eq.~\eqref{Bn_def}, we can deduce that
$\mu(B_n) \leq d2^{-6(n-1)}$ for all $n \geq 0$. Taking $\epsilon\rightarrow0^+$ in Eq.~\eqref{W2_bound}, we conclude that:
\begin{equation}
    W_2^2(\mu, \nu)\leq 2\mathcal{D}_2(\mu, \nu).
    \label{wd_bound}
\end{equation}

We will need the following lemma to estimate $\mathcal{D}_2(\mu, \nu)$.

\begin{lemma}
\rm
\label{appendix_lemma2}
(\cite[Lemma 6]{fournier2015rate}) Let $d \geq 1$. For all $\mu, \nu \in \mathcal{P}(\mathbb{R}^d)$,
\begin{equation}
    \mathcal{D}_2(\mu, \nu) \leq \frac{5}{3} \sum_{n\geq 0} 2^{2n} \sum_{\ell\geq 0} 2^{-2\ell} \sum_{F\in\mathcal{P}_\ell} \left| \mu(2^n F \cap B_n) - \nu(2^n F \cap B_n) \right|,
\end{equation}
where $\mathcal{D}_2(\mu, \nu)$ is defined in Eq.~\eqref{D_def},
and $2^n F \coloneqq \{2^n x : x \in F\}$.
\end{lemma}

We assume that $\mu \in \mathcal{P}(\mathbb{R}^d)$. 
From Eq.~\eqref{wd_bound}, we only need an upper bound for $\mathbb{E}[\mathcal{D}_2(\mu_N,\mu)]$. In the following, the positive constant $C$, 
whose value may change, depends only on $d$.
For a Borel subset $A \subset \mathbb{R}^d$, since $N\mu_N(A)$ is $\mathrm{Binomial}(N,\mu(A))$-distributed, we 
have
\[
\mathbb{E}\big[|\mu_N(A) - \mu(A)|\big] \leq \min\left\{2\mu(A),\sqrt{\mu(A)/N}\right\}.
\]
We denote the number of elements in $\mathcal{P}_{\ell}$ by $|\mathcal{P}_\ell| \coloneqq s_{\ell}2^{d(\ell+1)}$.
Using the Cauchy-Schwarz inequality, we deduce that for all 
$n \geq 0$, all $\ell \geq 0$,
\[
\sum_{F\in\mathcal{P}_\ell} \mathbb{E}\left[|\mu_N(2^n F \cap B_n) - \mu(2^n F \cap B_n)|\right] \leq \min\left\{2\mu(B_n), s_{\ell}^{\frac{1}{2}}2^{\frac{d(\ell+1)}{2}}(\mu(B_n)/N)^{\frac{1}{2}}\right\}.
\]
Using Lemma~\ref{appendix_lemma2} and the fact that $\mu(B_n) \leq d2^{-q(n-1)}$, we have:
\begin{equation}
\begin{aligned}
\mathbb{E}[\mathcal{D}_2(\mu_N,\mu)] &\leq C\sum_{n\geq 0} 2^{2n}\sum_{\ell\geq 0} 2^{-2\ell} \min\left\{2^{-6n}, s_{\ell}^{\frac{1}{2}}2^{\frac{d(\ell+1)}{2}}(2^{-6n}/N)^{\frac{1}{2}}\right\}.\\
&\leq C \sum_{\ell\geq 0}2^{-2\ell}\min\left\{1, s_{\ell}^{\frac{1}{2}}2^{\frac{d(\ell+1)}{2}}N^{-\frac{1}{2}}\right\}.
\end{aligned}
\label{estimates}
\end{equation}

Specifically, in Eq.~\eqref{estimates}, there exists an $\ell_0$ such that when $\ell\geq\ell_0$, 
\begin{equation}
    s_{\ell}^{\frac{1}{2}}\leq 2s,
\end{equation}
where $s\coloneqq \big(\prod_{i=1}^d\frac{\sigma_i}{\sigma_1}\big)^{\frac{1}{2}}$ (in $\mathcal{P}_{\ell}$, for the $i^{\text{th}}$ dimension, the number of translations of the interval $(0, \frac{1}{2^{\ell}}]$ needed is $\lceil 2^{\ell+1}\sigma_i \rceil\rightarrow2^{\ell+1}\sigma_i$ as $\ell\rightarrow\infty$). We denote:
\begin{equation}
    \ell_1=\Big[\frac{2}{d}\log_2 \frac{N^{\frac{1}{2}}}{2s} \Big]-1.
\end{equation}
When $N$ is sufficiently large, $\ell_1>>\ell_0$. Therefore, we have:
\begin{equation}
\begin{aligned}
    \mathbb{E}\big[\mathcal{D}_2(\mu_N,\mu)\big]&\leq C\Big(\sum_{\ell\geq0}^{\ell_0} 2^{\frac{(d-4)(\ell+1)}{2}}N^{-\frac{1}{2}}+ \sum_{\ell\geq \ell_1+1}2^{-2(\ell+1)} +\sum_{\ell\geq\ell_0+1}^{\ell_1} 2s2^{\frac{(d-4)(\ell+1)}{2}}N^{-\frac{1}{2}} \Big)\\
    &\leq 2C\big(2^{\frac{(d-4)\ell_0}{2}}N^{-\frac{1}{2}} + 2^{-2\ell_1}\big)\\
    &\leq 2C\Big(2^{\frac{(d-4)\ell_0}{2}}N^{-\frac{1}{2}} + \big(\frac{N}{(2s)^2}\big)^{-\frac{2}{d}}\Big)\\
    &\leq 2CM^2\Big(N^{-\frac{1}{2}} + \big(\prod_{i=1}^d(\frac{\sigma_i}{\sigma_1})^2N\big)^{-\frac{2}{d}}\Big),
    \end{aligned}
    \label{proofa_result}
\end{equation}
which proves Theorem~\ref{theorem_2}. When $\sigma_1\neq 1$, we denote $\tilde{\bm{y}}\coloneqq \frac{\bm{y}}{\sigma_1}$ with an associate probability measure $\tilde{\mu}$ and $N$-sample empirical probability measure $\tilde{\mu}_N$, respectively. Then, using the second last step in Eq.~\eqref{proofa_result}, we have:
\begin{equation}
    \E[W_2^2(\mu_N, \mu)] = \sigma_1^2\E[W_2^2(\tilde{\mu}_N, \tilde{\mu})]\leq 2CM^2\Big(2^{(d-4)\ell_0/2}N^{-\frac{1}{2}} + \big(\frac{N}{(2s)^2}\big)^{-\frac{2}{d}}\Big),
\end{equation}
completing the proof of Theorem~\ref{theorem_2}. 

\section{Training settings and hyperparameters}
\label{training_details}
{\scriptsize \begin{table}[h!]
\centering
\caption{\footnotesize Training hyperparameters, hyperparameters in the neural network model, and training settings for each example. The neural network parameters include means and standard deviations $a_{i, j, k}, \sigma_{i, j, k}$ for weights $w_{i, j, k}$, the weights $\tilde{w}_{i, j, k}$ (for the ResNet technique), as well as biases $b_{i, k}$ in Fig.~\ref{fig:snn}.} 
{\scriptsize\begin{tabular}{lllll}
\toprule
 & Example~\ref{example1} & Example~\ref{example3_case} \\
\midrule
gradient descent method & Adam & Adam  \\
forward propagation method &  ResNet  & Normal\\
learning rate & 0.005  & 0.005  \\
number of epochs &1000  & 400   \\
number of training samples & 4000  & 300  \\
size of neighborhood $\delta$ in the loss Eq.~\eqref{updated_loss} & 0.25&  0.125\\
number of hidden layers in $\Theta$ & 4  &2 \\
activation function &  ELU  &ELU \\
number of neurons in each layer & 40  &60  \\
initialization for $w_{i, j, k}$ and $b_{i, j}$  & $\mathcal{N}(0, 0.01^2)$  & $\mathcal{N}(0, 0.01^2)$\\
initialization for $\tilde{w}_{i, j, k}$  & \texttt{torch.nn} default   &$\backslash$\\
\bottomrule
\end{tabular}}
\label{tab:setting}
\end{table}}

\section{Robustness of the distribution of the output from the SNN model in Fig.~\ref{fig:snn} against perturbations in its parameters}
\label{snn_robust}
For a fixed input $\bm{x}$ into the SNN model in Fig.~\ref{fig:snn}, the probability measure of the output is ``continuously dependent" on the biases and means as well as variances of the weights under the $W_2$ metric. We prove the following result.
\begin{theorem}
\rm
    Suppose two SNNs in Fig.~\ref{fig:snn}, denoted by $S_1$ and $S_2$, have the same structure, \textit{i.e.}, the same number of hidden layers $L$ and the same number of neurons $H_1,...,H_L$ in the layer $1,..,L$, respectively. We denote the biases and means as well as variances for weights in $S_1$ to be $\{b_{i, k}\}$, $\{a_{i, j, k}\}$, and $\{\sigma_{i, j, k}\}$; we denote the biases and means as well as variances for weights in $S_2$ to be $\{\hat b_{i, k}\}$, $\{\hat a_{i, j, k}\}$, and $\{\hat \sigma_{i, j, k}\}$.
    Furthermore, the \texttt{ReLU} function is used as the activation function. Given $\bm{x}\in\mathbb{R}^d$, we denote the probability measures of the outputs of $S_1$ and $S_2$ by $\mu_{\bm{x}, 1}$ and $\mu_{\bm{x}, 2}$, then we have:
    \begin{equation}
    \begin{aligned}
                W_2^2(\mu_{\bm{x}, 1}, \mu_{\bm{x}, 2})&\leq C\big(\sum_{i=0}^L\sum_{j=1}^{H_i}\sum_{k=1}^{H_{i+1}}(a_{i, j, k}-a_{i, j, k})^2\\
    &\quad\quad\quad+\sum_{i=0}^L\sum_{j=1}^{H_i}\sum_{k=1}^{H_{i+1}}(\sigma_{i, j, k}-\sigma_{i, j, k})^2 + \sum_{i=0}^L\sum_{j=1}^{H_{i+1}}(b_{i, j}-\hat{b}_{i, j})^2 \big),
    \end{aligned}
        \label{robust_bound}
    \end{equation}
    where $C$ is a constant that depends on $\bm{x}$ as well as $\{ a_{i, j, k}\}$, $\{ \sigma_{i, j, k}\}$, $\{\hat a_{i, j, k}\}$, $\{\hat \sigma_{i, j, k}\}$, $\{b_{i, j}\}$, and $\{\hat b_{i, j}\}$.
\end{theorem}

\begin{proof}
For the $i^{\text{th}}$ layer of both SNNs, we denote their outputs to be  
 $\bm{x}^{i, 1}\coloneqq (x_1^{i, 1},...,x_{H_i}^{i, 1})$ and $\bm{x}^{i, 2}\coloneqq (x_1^{i, 1},...,x_{H_i}^{i, 1})$ with associated probability measures $\mu^{i, 1}$ and $\mu^{i, 2}$, respectively (when $i=0$, $\bm{x}^{0, 1}=\bm{x}^{0, 2}=\bm{x}$). Then, for any $\epsilon>0$, there exists a coupling measure $\pi$ for $(\bm{x}^{i, 1}, \bm{x}^{i, 2})$ such that:
\begin{equation}
    \E_{(\bm{x}^{i, 1},\bm{x}^{i, 2})\sim \pi}\big[\|\bm{x}-\bm{y}\|^2\big]\leq W_2^2(\mu^{i, 1}, \mu^{i, 2})+\epsilon.
\end{equation}
The input of the $(i+1)^{\text{th}}$ layer will then be:
\begin{equation}
\begin{aligned}
    &\bm{x}^{i+1, 1}\coloneqq ({x}_1^{i+1, 1},...,{x}^{i+1, 1}_{H_{i+1}}), \,\,{\bm{x}}^{i+1, 2}\coloneq({x}_1^{i+1, 1},...,x_{H_{i+1}}^{i+1, 1}),\\
    &{x}_j^{i+1, 1}\coloneqq\sum_{k=1}^{H_i} w_{i, j, k}x_k^{i, 1},\,\, {x}_j^{i+1, 2}\coloneqq\sum_{k=1}^{H_i} \hat{w}_{i, j, k}x_k^{i, 2}.
    \end{aligned}
\end{equation}
For every pair of $(j, k), j=1,...,H_{i+1}, k=1,...,H_i$, since both weights $w_{i, j, k}$ and $\hat{w}_{i, j, k}$ obey the normal distribution, there exists a coupling probability measure of $(w_{i, j, k}, \hat{w}_{i, j, k})$, denoted by $\pi_{i, j, k}$, satisfying:
\begin{equation}
    \E_{(w_{i, j, k}, \hat{w}_{i, j, k})\sim\pi_{i, j, k}}\big[\|w_{i, j, k}-\hat{w}_{i, j, k}\|^2\big]\leq (a_{i, j, k} - \hat{a}_{i, j, k})^2+(\sigma_{i, j, k}-\hat{\sigma}_{i, j, k})^2.
\end{equation}
Furthermore, the marginal distributions of $\pi_{i, j, k}$ coincide with the probability distributions associated with $w_{i, j, k}$ and $\hat{w}_{i, j, k}$, respectively. Since the independently sampled weights $\{w_{i, j, k}\}$, $\{\hat{w}_{i, j, k}\}$ are independent of $\bm{x}^{i,1}$ and $\bm{x}^{i,2}$, respectively, we have:
\begin{equation}
\begin{aligned}
    &W_2^2(\mu^{i+1, 1}, \mu^{i+1, 2})\leq \E_{(\bm{x}^{i, 1},\bm{x}^{i, 2})\sim\pi, (w_{i, j, k},\hat{w}_{i, j, k})\sim\pi_{i, j, k}}\big[\|{\bm{x}}^{i+1,1}-{\bm{x}}^{i+1,2}\|^2\big]\\
    &\hspace{2.6cm}= \E_{(\bm{x}^{i, 1},\bm{x}^{i, 2})\sim\pi, (w_{i, j, k},\hat{w}_{i, j, k})\sim\pi_{i, j, k}}\big[\sum_{j=1}^{H_{i+1}}({x}^{i+1,1}_j-{x}^{i+1,2}_j)^2\big]\\
    &\hspace{1.5cm}\leq H_i\sum_{j=1}^{H_{i+1}}\sum_{k=1}^{H_i}\E_{(\bm{x}^{i, 1},\bm{x}^{i, 2})\sim\pi, (w_{i, j, k},\hat{w}_{i, j, k})\sim\pi_{i, j, k}}[(w_{i, j, k}x_k^{i, 1}-\hat{w}_{i, j, k}x_k^{i, 2})^2] \\
    &\hspace{2.2cm}+ \sum_{j=1}^{H_{i+1}}(b_{{i+1}, j}-\hat{b}_{{i+1}, j})^2\\
    &\hspace{1.2cm}\leq 2H_i\sum_{j=1}^{H_{i+1}}\sum_{k=1}^{H_i}\E_{(\bm{x}^{i, 1},\bm{x}^{i, 2})\sim\pi, (w_{i, j, k},\hat{w}_{i, j, k})\sim\pi_{i, j, k}}[(w_{i, j, k}x_k^{i, 1}-\hat{w}_{i, j, k}x_k^{i, 1})^2] \\
&\hspace{1.4cm}+2H_i\sum_{j=1}^{H_{i+1}}\sum_{k=1}^{H_i}\E_{(\bm{x}^{i, 1},\bm{x}^{i, 2})\sim\pi, (w_{i, j, k},\hat{w}_{i, j, k})\sim\pi_{i, j, k}}[(\hat{w}_{i, j, k}x_k^{i, 1}-\hat{w}_{i, j, k}x_k^{i, 2})^2]\\
&\hspace{2.2cm}+ \sum_{j=1}^{H_{i+1}}(b_{{i+1}, j}-\hat{b}_{{i+1}, j})^2\\
    &\hspace{2cm}\leq 2H_i\sum_{j=1}^{H_{i+1}}\sum_{k=1}^{H_i}\E_{(\bm{x}^{i, 1},\bm{x}^{i, 2})\sim\pi, (w_{i, j, k},\hat{w}_{i, j, k})\sim\pi_{i, j, k}}[(\hat{w}_{i, j, k}x_k^{i, 1}-\hat{w}_{i, j, k}x_k^{i, 2})^2]\\
&\hspace{-0.5cm}+2H_i\sum_{j=1}^{H_{i+1}}\sum_{k=1}^{H_i}\E[(x_j^{i, 1})^2]\cdot\big((a_{i, j, k}-\hat{a}_{i, j, k})^2+(\sigma_{i, j, k}-\hat{\sigma}_{i, j, k})^2\big) + \sum_{j=1}^{H_{i+1}}(b_{i+1, j}-\hat{b}_{i+1, j})^2\\
&\hspace{2cm}\leq 2H_i \max_{k=1,...,H_{i}} \big(\sum_{j=1}^{H_{i+1}} (\hat{a}_{i, j, k}^2+\sigma_{i, j, k}^2)\big)\E_{(\bm{x}^{i, 1},\bm{x}^{i, 2})\sim\pi}\big[\sum_{k=1}^{H_i}\|x_k^{i, 1} - x_k^{i, 2}\|^2\big]\\
&\hspace{-0.5cm}+2H_i\sum_{j=1}^{H_{i+1}}\sum_{k=1}^{H_i}\E[(x_k^{i, 1})^2]\cdot\big((a_{i, j, k}-\hat{a}_{i, j, k})^2+(\sigma_{i, j, k}-\hat{\sigma}_{i, j, k})^2\big) + \sum_{j=1}^{H_{i+1}}(b_{{i+1}, j}-\hat{b}_{{i+1}, j})^2\\
&\hspace{2cm}\leq 2H_i \max_{k=1,...,H_{i}} (\sum_{k=1}^{H_i} \hat{a}_{i, j, k}^2+\hat{\sigma}_{i, j, k}^2)(W_2^2(\mu^{i, 1}, \mu^{i, 2})+\epsilon) \\
&\hspace{-0.5cm}+2H_i\sum_{j=1}^{H_{i+1}}\sum_{k=1}^{H_i}\E[(x_k^{i, 1})^2]\cdot\big((a_{i, j, k}-\hat{a}_{i, j, k})^2+(\sigma_{i, j, k}-\hat{\sigma}_{i, j, k})^2\big) + \sum_{j=1}^{H_{i+1}}(b_{{i+1}, j}-\hat{b}_{{i+1}, j})^2.
    \end{aligned}
\end{equation}
Taking $\epsilon\rightarrow 0$, we conclude that:
\begin{equation}
\begin{aligned}
        W_2^2({\mu}^{i+1, 1}, \mu^{i+1, 2})\leq &C_0\Big(W_2^2(\mu^{i, 1}, \mu^{i, 2}) + \sum_{j=1}^{H_{i+1}}\sum_{k=1}^{H_i}\big((a_{i, j, k}-\hat{a}_{i, j, k})^2\\
        &\hspace{0.5cm}+(\sigma_{i, j, k}-\hat{\sigma}_{i, j, k})^2\big)+ \sum_{j=1}^{H_{i+1}}(b_{{i+1}, j}-\hat{b}_{{i+1}, j})^2\Big),
\end{aligned}
    \label{iterate}
\end{equation}
where $C_0$ depends on $\hat{a}_{i, j, k}, \hat{\sigma}_{i, j, k}, j=1,..,H_{i+1}, k=1,..., H_i$, and $\E[(x_k^{i, 1})^2], k=1...,H_i$.

Note that $\|\text{ReLU}(\bm{x}^{{i+1}, 1})-\text{ReLU}({\bm{x}}^{{i+1}, 2})\|\leq \|{\bm{x}}^{{i+1}, 1}-{\bm{x}}^{{i+1}, 2}\|$. Furthermore, for any coupling probability measure of $(\bm{x}^{{i+1}, 1}, \bm{x}^{{i+1}, 2})$ whose marginal distributions coincide with $\mu^{i+1, 1}$ and $\mu^{i+1, 2}$, if we denote the 
We denote $\tilde \mu^{H_{i+1}, 1}$ and $\tilde \mu^{H_{i+1}, 2}$ to be the probability measures associated with $\text{ReLU}({\bm{x}^{{i+1}, 1}})$ and $\text{ReLU}({\bm{x}^{{i+1}, 2}})$. For any coupling probability measure $\pi$ of $(\bm{x}^{{i+1}, 1}, \bm{x}^{{i+1}, 2})$ whose marginal distributions coincide with $\mu^{i+1, 1}$ and $\mu^{i+1, 2}$, respectively, if we denote $\tilde{\pi}$ to be the pushforward probability measure of $(\text{ReLU}(\bm{x}^{{i+1}, 1}), \text{ReLU}(\bm{x}^{{i+1}, 2}))$ under $\pi$, it can be easily observed that the marginal distributions of $\tilde{\pi}$ coincide with $\tilde \mu^{H_{i+1}, 1}$ and $\tilde \mu^{H_{i+1}, 2}$, respectively. Therefore, we have:
\begin{equation}
    W_2^2(\tilde \mu^{{i+1}, 1}, \tilde \mu^{{i+1}, 2})\leq W_2^2(\mu^{{i+1}, 1}, \mu^{{i+1}, 2}).
    \label{relu_bound}
\end{equation}
Since the input for both $S_1$ and $S_2$ is the same $\bm{x}$, iterating the inequalities Eqs.~\eqref{iterate} and~\eqref{relu_bound} over all hidden layers, we conclude that:
\begin{equation}
\begin{aligned}
        W_2^2(\mu_{\bm{x}, 1}, \mu_{\bm{x}. 2})&\leq C\big(W_2^2(\delta(\tilde{\bm{x}}-\bm{x}), \delta(\tilde{\bm{x}}-\bm{x})) + \sum_{i=0}^L\sum_{j=1}^{H_{i+1}}\sum_{k=1}^{H_{i}}\big((a_{i, j, k}-\hat{a}_{i, j, k})^2\\
    &\hspace{1cm}+(\sigma_{i, j, k}-\hat{\sigma}_{i, j, k})^2\big) + \sum_{i=0}^L\sum_{j=1}^{H_{i+1}}(b_{i+1, j}-\hat{b}_{i+1, j})^2\big), 
\end{aligned}
\label{thm2_result}
\end{equation}
where $\delta(\tilde{\bm{x}}-\bm{x})$ is the Dirac delta measure for the random variable $\tilde{\bm{x}}\in D$, and $C$ depends on $\bm{x}$ as well as $\{ a_{i, j, k}\}$, $\{ \sigma_{i, j, k}\}$, $\{\hat a_{i, j, k}\}$, $\{\hat \sigma_{i, j, k}\}$, $\{b_{i, j}\}$, and $\{\hat b_{i, j}\}$. Noting that the first term on the RHS of Eq.~\eqref{thm2_result} is 0, we have proved Eq.~\eqref{robust_bound}. 

\end{proof}

\bibliographystyle{siamplain}
\bibliography{reference}
\end{document}